%% file: elsarticle-template.tex
\documentclass[preprint,12pt,numbers,sort&compress]{elsarticle}

\usepackage{lineno}

\usepackage{amssymb}            
\usepackage{amsmath}            
\usepackage{amsthm}             

\usepackage{graphicx}           
\usepackage{placeins}
\usepackage{subcaption}         
\usepackage{wrapfig}            
\usepackage{float}              
\usepackage{stfloats}           
\usepackage{caption}            

\usepackage{array}              
\usepackage{booktabs}           
\usepackage{tabularx}           
\usepackage{longtable}          
\usepackage{multirow}           
\usepackage{makecell}           
\usepackage{colortbl}           

\usepackage{algorithm}          
\usepackage{algpseudocode}      
\usepackage{listings}           

\usepackage{bm}                 
\usepackage{mathtools}          
\usepackage{pifont}             
\usepackage{setspace}           

\usepackage{enumitem}           
\usepackage{hyperref}           
\usepackage[accsupp]{axessibility} 
\usepackage{color, xcolor}      
\usepackage{url}                

\usepackage{lineno}             
\usepackage{xparse}
\usepackage{xcolor}
\usepackage{hyperref}
\journal{CMPB}

\begin{document}

\begin{frontmatter}



\title{MMLNB: Multi-Modal Learning for Neuroblastoma Subtyping Classification Assisted with Textual Description Generation} 

\author[label1]{Huangwei Chen\corref{cofirst}}
\ead{chenhuangwei@hdu.edu.cn}
\author[label2]{Yifei Chen\corref{cofirst}}
\ead{justlfc03@gmail.com}
\author[label1]{Zhenyu Yan}
\ead{22050737@hdu.edu.cn}
\author[label1]{Mingyang Ding}
\ead{dingmingyang@hdu.edu.cn}
\author[label1]{Chenlei Li}
\ead{lichenlei@hdu.edu.cn}
\author[label3,label4]{Zhu Zhu\corref{mycorrespondingauthor}}
\author[label1]{Feiwei Qin\corref{mycorrespondingauthor}}

\cortext[cofirst]{These authors contributed equally to this work.}
\cortext[mycorrespondingauthor]{Corresponding author. E-mail address: \href{zhuzhu_cs@zju.edu.cn}{zhuzhu\_cs@zju.edu.cn}, \href{qinfeiwei@hdu.edu.cn}{qinfeiwei@hdu.edu.cn}}

\affiliation[label1]{organization={Hangzhou Dianzi University},
            city={Hangzhou},
            postcode={310018}, 
            state={Zhejiang},
            country={China}}

\affiliation[label2]{organization={Tsinghua University},
            postcode={100084}, 
            state={Beijing},
            country={China}}

\affiliation[label3]{organization={National Clinical Research Center for Child Health, National Children's Regional Medical Center, Children's Hospital, Zhejiang University School of Medicine}, 
            city={Hangzhou}, 
            postcode={310052}, 
            state={Zhejiang}, 
            country={China}}
            
\affiliation[label4]{organization={Sino-Finland Joint AI Laboratory for Child Health of Zhejiang Province}, 
            city={Hangzhou}, 
            postcode={310052}, 
            state={Zhejiang}, 
            country={China}}

\begin{abstract}
Neuroblastoma (NB), a leading cause of childhood cancer mortality, exhibits significant histopathological variability, necessitating precise subtyping for accurate prognosis and treatment. Traditional diagnostic methods rely on subjective evaluations that are time-consuming and inconsistent. To address these challenges, we introduce MMLNB, a multi-modal learning (MML) model that integrates pathological images with generated textual descriptions to improve classification accuracy and interpretability. The approach follows a two-stage process. First, we fine-tune a Vision-Language Model (VLM) to enhance pathology-aware text generation. Second, the fine-tuned VLM generates textual descriptions, using a dual-branch architecture to independently extract visual and textual features. These features are fused via Progressive Robust Multi-Modal Fusion (PRMF) Block for stable training. Experimental results show that the MMLNB model is more accurate than the single modal model. Ablation studies demonstrate the importance of multi-modal fusion, fine-tuning, and the PRMF mechanism. This research creates a scalable AI-driven framework for digital pathology, enhancing reliability and interpretability in NB subtyping classification. Our source code is available at \href{https://github.com/HovChen/MMLNB}{https://github.com/HovChen/MMLNB}.
\end{abstract}




\end{frontmatter}



\input{sections/1_introduction}

\input{sections/2_related_work}

\input{sections/3_methods}

\input{sections/4_experiment}

\input{sections/5_conclusion}

\section*{Compliance with Ethical Standards}
The study was conducted in accordance with the Declaration of Helsinki (as revised in 2013). The study was approved by the Academic Ethics Committee of Children’s Hospital Affiliated to Zhejiang University School of Medicine (No. 2023-IRB-0287-P-01). Individual consent for this retrospective analysis was waived.

\section*{Acknowledgements}
The authors acknowledge the financial support of the Medical Health Science and Technology Project of Zhejiang Provincial Health Commission (2023KY832).

\clearpage
\bibliographystyle{elsarticle-num-names}

\end{document}

%% file: sections/1_introduction.tex
\section{Introduction}
\label{sec1}
Neuroblastoma (NB) is one of the most common solid extracranial tumors in children, accounting for approximately 15\% of pediatric cancer-related deaths. Originating from neural crest cells, it primarily affects the adrenal glands but can also arise in other tissues of the sympathetic nervous system. The disease exhibits significant heterogeneity, ranging from spontaneously regressing tumors to highly aggressive cases with widespread metastases. Given this variability, an accurate classification of NB is crucial for the prognosis and treatment planning. However, current histopathological classification methods rely on expert-driven morphological assessments, which can be subjective and prone to differences among observers~\cite{chen2024sckansformer}.

Histopathological examination remains the gold standard for NB diagnosis and subtyping, which includes assessing cellular composition, differentiation, and mitosis-karyorrhexis index (MKI). The International NB Pathology Classification (INPC) system stratifies tumors on differentiation status, MKI, and patient age, with well-differentiated, low-MKI tumors linked to a better prognosis, while undifferentiated, high-MKI tumors indicate aggressive disease. However, tumor heterogeneity complicates classification, as varying differentiation levels within the same tumor create challenges. Inter-observer variability further affects consistency, particularly in borderline cases, influencing risk stratification and treatment. Manual histopathological assessment is also time consuming, causing bottlenecks in high-volume clinical settings and posing challenges in regions with limited pathological expertise. These limitations highlight the need for standardized and scalable solutions to improve precision, efficiency, and accessibility in NB diagnosis~\cite{chen2024accurate}.

Recent advancements in artificial intelligence (AI) and deep learning have significantly improved medical image analysis, particularly in digital pathology. Convolutional neural networks (CNN) have been widely employed for extracting and classifying histopathological features, while natural language processing (NLP) models have shown promise in processing and structuring pathology reports. However, single-modal AI models remain inherently limited—image-based models lack the contextual understanding provided by textual reports, while text-based models cannot independently verify findings without visual evidence. This disconnect reduces interpretability and generalizability, limiting the reliability of AI-driven diagnostic tools~\cite{zhan2024medm2g}.

To address the challenges in NB pathology subtyping, we propose a dual-branch model that integrates deep learning-based image analysis with generated textual descriptions to improve diagnostic accuracy and model interpretability. Specifically, we fine-tuned Qwen2.5-VL-7B-Instruct, a Vision-Language Model (VLM) developed by Alibaba Cloud~\cite{bai2025qwen2}, using our NBITP-1.5K dataset, which includes 1,500 pathological image-text pairs. This process enables the model to learn complex visual-textual correlations. In our dual-branch feature extraction strategy, a VGG16-based~\cite{simonyan2014very} CNN extracts visual features from pathological images, while the fine-tuned VLM generates pathological descriptions from these images. The generated textual descriptions are then encoded using Bidirectional Encoder Representations from Transformers (BERT)~\cite{devlin2019bert} to extract textual embeddings. We then merge these textual embeddings with the visual embeddings from VGG16 through our proposed Progressive Robust Multimodal Fusion (PRMF) Block. This block dynamically adjusts the weights of both modalities based on text confidence, ensuring smooth training and producing a unified multimodal representation for the final subtyping classification.

The key contributions of this paper include:
\begin{itemize}
    \item \textbf{Fine-Tuned VLM for Pathology-Aware Text Generation:} Our customized fine-tuning of the VLM enables high-fidelity pathology-aware text generation, converting abstract feature representations into clinically meaningful descriptions. This not only improves interpretability, but also provides expert-level diagnostic insights, effectively bridging the gap between deep learning-based image analysis and human-readable pathology reports.
    
    \item \textbf{Text Noise-Robust Cross-Modal Feature Fusion:} We have implemented a robust feature fusion mechanism that effectively mitigates the interference of text noise in results through a text confidence evaluation strategy, effectively leveraging complementary cross-modal information. This approach improves classification accuracy, ensures more comprehensive characterization of diseases, and greatly enhances interoperability in multimodal AI frameworks.
    
    \item \textbf{Scalable Multi-Modal AI Framework for Digital Pathology:} Our approach creates a solid foundation for a wider range of multimodal AI applications in medical imaging and digital pathology. This development leads to automated, scalable diagnostic tools that improve clinical decision making and improve patient outcomes.
\end{itemize}

%% file: sections/2_related_work.tex
\section{Related Work}
\subsection{Traditional NB Diagnostic Approaches}
NB, the most common solid extracranial tumor in children and the ``King of Pediatric Tumors'', accounts for > 15\% of cancer deaths in children~\cite{yue2025global}. Over 50\% of cases occur in children under two years old~\cite{fredlund2024moxd1}, and its clinical presentation is a significant heterogeneity~\cite{chen2024developmental}, ranging from asymptomatic masses to life-threatening metastatic disease~\cite{nakazawa2021biological}. The International NB Pathology Classification (INPC) remains the gold standard for histopathological evaluation, categorizing tumors into three subtypes, namely undifferentiated (UD), poorly differentiated (PD), and differentiated (D) subtype, based on the grade of differentiation~\cite{asgharzadeh2024neuroblastoma}.

Traditional diagnostics in pediatrics heavily depend on manual microscopic examinations, which must be conducted by skilled pediatric pathologists. This demanding method is subject to variability among different observers~\cite{YANG2023107583}. This issue is further intensified by the global shortage of pediatric oncology specialists, particularly in resource-limited regions, leading to delays or inaccuracies in diagnoses~\cite{zhu2025global}. This scenario underscores the urgent need for automated solutions to enhance clinical workflows and improve access to accurate diagnostic expertise.

\subsection{Deep Learning Approaches in Histopathological Classification}
Recent advances in deep learning have demonstrated promise in automating NB classification. Early studies, such as the work by~\cite{gheisari2018convolutional}, employed convolutional deep belief networks (CDBNs) to classify 1,043 NB histological images into five subtypes (UD, PD, D, ganglioneuroblastoma (GNB), and ganglioneuroma (GN)), achieving improved accuracy over handcrafted features. Subsequent approaches integrated scale-invariant feature transforms (SIFT) with support vector machines (SVMs) to encode texture patterns, yet these methods struggled with tumor heterogeneity and granularity shifts~\cite{gheisari2018computer}.

Modern frameworks like SuperHistopath~\cite{zormpas2021superhistopath} and ScoreNet~\cite{stegmuller2023scorenet} leveraged CNNs and transformer-based attention mechanisms to enhance tumor heterogeneity mapping. However, single-modality models remain limited by interpretability, hindering clinical trust. Although methods like HiFuse~\cite{huo2024hifuse} and FMDNN~\cite{ding2024fmdnn} introduced multiscale fuzzy logic-based reasoning, they failed to resolve the ``black-box'' nature of deep learning predictions, particularly in distinguishing high-risk tumor subtypes~\cite{shimada2020neuroblastoma}. To overcome these challenges, multimodal learning (MML) offers a promising path to improving both accuracy and interpretability~\cite{chen2022pan}.

\subsection{Multi-Modal Learning for Histopathological Classification}
Recent studies~\cite{li2023survival, mahootiha2024multimodal,chen2025toward} have demonstrated the potential of multimodal learning in biomedical applications. However, these methods largely rely on preexisting paired modalities, limiting their applicability when such paired data is scarce. Specifically, in histopathology, the lack of paired textual descriptions and pathological images poses a significant challenge for MML. Moreover, in recent years, VLMs such as InternVL~\cite{chen2024internvlscalingvisionfoundation} , MiniCPM-V~\cite{hu2024minicpmunveilingpotentialsmall} and Qwen-VL~\cite{bai2025qwen2} have achieved remarkable progress in open-domain tasks by leveraging large-scale internet data for pretraining. These advancements have significantly improved image-based text generation capacity. However, existing VLMs are optimized primarily for natural image semantics~\cite{zhang2024vlm} rather than domain-specific pathological features. Consequently, their sensitivity to fine-grained clinical patterns remains suboptimal.

To bridge this gap, several studies have explored domain adaptation techniques to enhance the applicability of VLMs in medical imaging. For instance, Eslami et al.~\cite{eslami2021doesclipbenefitvisual} fine-tuned biomedical VLMs on radiology reports to improve medical description generation. Lu et al.~\cite{Lu2024} introduced PathChat, a model incorporating pathological prior knowledge to refine multimodal pathology dialogues. These studies highlight the importance of task-specific fine-tuning and domain-specific knowledge integration in improving the performance of VLMs for medical applications. However, most of the existing research has focused on common adult diseases~\cite{kludt2024next,kondepudi2024foundation}, whereas pediatric-specific conditions such as NB remain largely unexplored. NB presents distinct imaging characteristics, including variations in primary lesion localization and microscopic-scale calcifications, necessitating the incorporation of pediatric anatomical priors. Moreover, the substantial distribution shift between general VLM pre-training datasets and specialized medical domains poses a significant challenge for direct adaptation.

To address these issues, we propose a novel method to generate pathological image text description using VLMs to assist pathological image classification. This method effectively introduces a missing modality and enhances the model's representation capacity. By doing so, our method not only mitigates the limitation of missing paired modalities but also improves the interpretability and robustness of histopathological classification.

%% file: sections/3_methods.tex
\section{Methods}
\subsection{Overall Architecture}
\begin{figure*}[ht]
\centering
\includegraphics[width=1.00\textwidth]{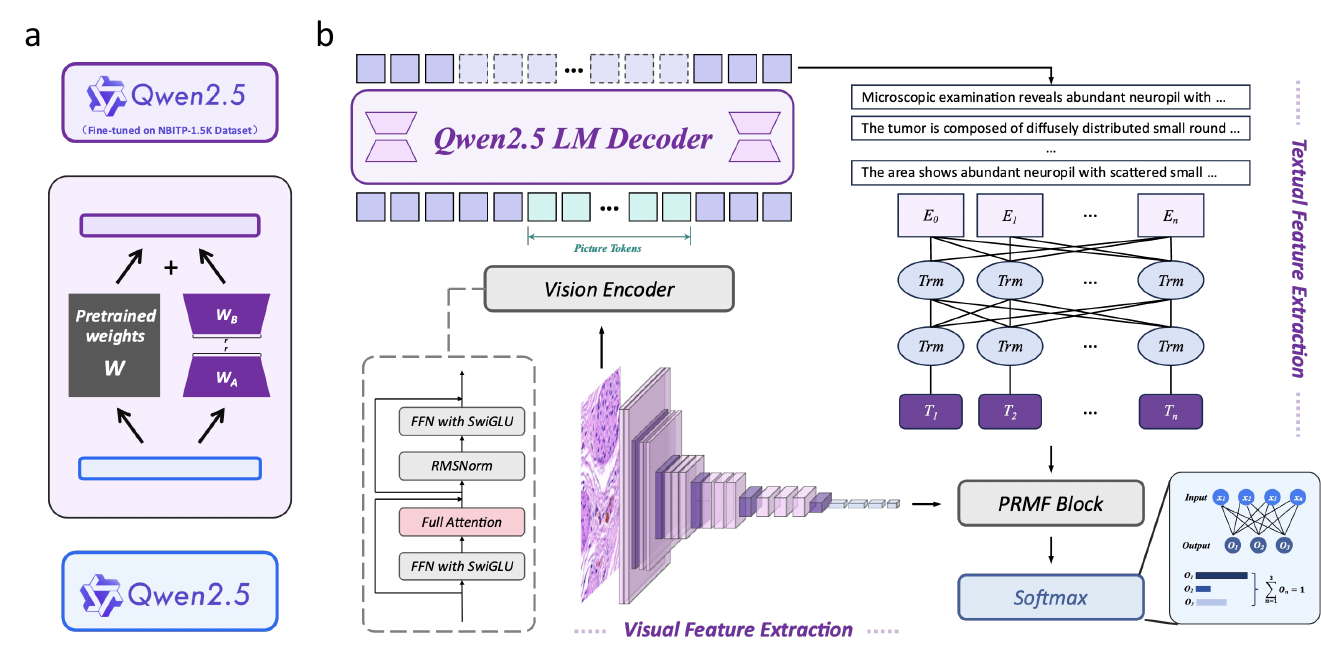}
\caption{The overall architecture of our proposed MMLNB model. The model consists of two stages: (a) fine-tuning a VLM via LoRA; (b) applying the fine-tuned VLM to generate pathology-aware textual descriptions to assist image classification.} 
\label{fig:pipeline}
\end{figure*}

Our proposed MMLNB model (Fig.~\ref{fig:pipeline}) follows a two-stage architecture, designed for efficient adaptation to NB pathology and accurate subtypes classification. In the first stage, we fine-tune the Qwen2.5-VL-7B-Instruct with Low-Rank Adaptation (LoRA), allowing it to learn domain-specific histopathological features. This adaptation is guided by expert-curated pathological image-text pairs, ensuring that the model effectively captures morphological characteristics and their corresponding textual descriptions. In the second stage, the fine-tuned VLM is integrated into a dual-branch classification framework. Here, a VGG16-based visual encoder extracts features from pathological images, while a BERT-based textual encoder processes textual descriptions generated by the VLM based on these pathological images. These features are then fused via the PRMF Block, which dynamically adjusts the weighting of image and text information based on a learnable confidence network. Finally, a classification module processes the fused features to predict the subtypes of NB, providing interpretable outputs that highlight key pathological features for improved diagnostic insights.

\subsection{NB-Specific LoRA Fine-Tuning}
While VLMs like Qwen2.5-VL demonstrate strong multimodal understanding across general domains, their adaptation to NB-specific tasks encounters three inherent limitations. First, there are domain-specific morphological patterns in NB imaging, such as variations in chromatin distribution, which fundamentally differ from the semantics of natural images. Second, there is a tendency for overfitting with full-parameter fine-tuning when there are limited NB image-text pairs. Lastly, there is a dual requirement to preserve general visual comprehension while acquiring precise domain-discriminative feature localization for pathological diagnosis.

To address these challenges, we employ LoRA for efficient parameter tuning in the domain adaptation of Qwen2.5-VL. This approach is grounded in the hypothesis that the adjustments made to the weight matrices during task adaptations lie within a low-rank manifold. By incorporating trained low-rank factorized matrices into pre-trained weight matrices, our approach facilitates the effective learning of specialized features while maintaining the integrity of the original knowledge. Formally, given a pre-training weight matrix $W_0 \in \mathbb{R}^{d \times k}$, the adaptation process is expressed as:

\begin{equation}
W = W_0 + \Delta W = W_0 + BA,
\end{equation}
where $B \in \mathbb{R}^{d \times r}$ and $A \in \mathbb{R}^{r \times k}$ are trainable parameter matrices, with the rank $r \ll \min(d,k)$ controlling adaptation complexity. This decomposition reduces parameter dimensionality from $d \times k$ to $r(d + k)$, achieving a substantial reduction in trainable parameters compared to full fine-tuning. This reduction is vital for reducing overfitting risks, especially in small-sample NB data scenarios. Theoretically, when task-specific updates lie in low-rank manifolds, LoRA constrains solution space while retaining model capacity.




Given the specialized demands of the NB understanding tasks, LoRA strategically enhances the mechanism of cross-modal interaction of Qwen2.5-VL. Let visual features be represented as \( v \in \mathbb{R}^{n_v \times d_v} \) (extracted from pathological images) and textual features as \( t \in \mathbb{R}^{n_t \times d_t} \) (extracted from generated textual descriptions). The cross-modal attention mechanism is improved through LoRA injection in its projection matrices:

\begin{equation}
(Q, K, V) = (W + B A) (v, t, t),
\end{equation}
where $W = (W_q, W_k, W_v) \in \mathbb{R}^{d \times d}$ represent frozen pretrained projection weights, and $B = (B_q, B_k, B_v)$, $A = (A_q, A_k, A_v)$ are trainable low-rank matrices with rank $r \ll d$. This hybrid parameterization strategy preserves the model's original cross-modal alignment capabilities while enabling targeted adaptation to pathological semantic relationships through:

\begin{equation}
\Delta W_{\mathit{cross}} = \bigcup_{m \in \{q,k,v\}} B_m A_m.
\end{equation}

The low-rank updates specifically improve the model's ability to form fine-grained associations between histopathological entities (e.g., nuclear morphology, tissue architecture patterns) and domain-specific terminologies.

\subsection{Multi-modal Feature Extraction}
\begin{figure}[ht]
\centering
\includegraphics[width=\columnwidth]{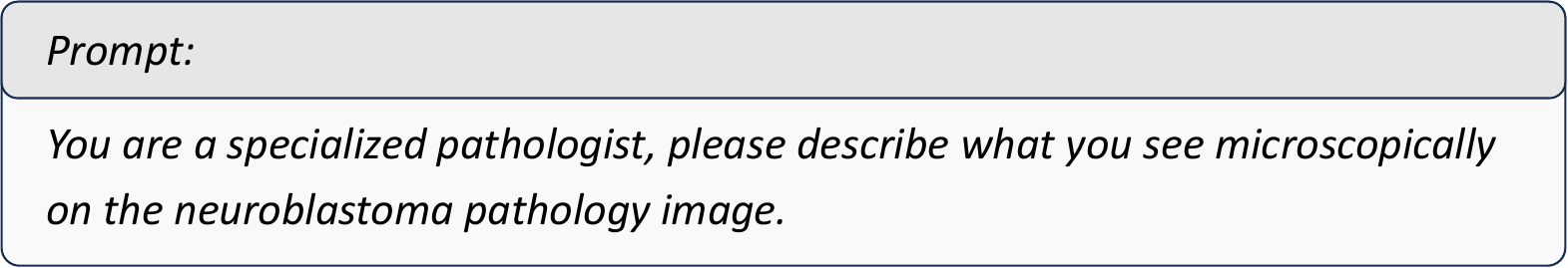}
\caption{The prompt for VLM. The structured prompt is designed to elicit precise pathological insights from Qwen2.5-VL.} 
\label{fig:prompt}
\end{figure}

Building on the LoRA fine-tuned Qwen2.5-VL, we employ a two-branch multimodal feature extraction framework to integrate histopathological imaging and textual descriptions for comprehensive NB analysis. This architecture consists of two branches: one directly processes pathological images through a visual encoder, while the other generates textual descriptions using the fine-tuned VLM, which are subsequently encoded by a text encoder for feature extraction. As illustrated in Fig.~\ref{fig:prompt}, the model is guided by a specialized prompt to ensure domain-relevant output.

The visual branch extracts deep hierarchical features using VGG16 archietechture~\cite{simonyan2014very}. In parallel, the textual branch employs BERT~\cite{devlin2019bert} for a context-rich text representation. This collaboration enables the model to establish robust correlations between morphological and textual features, enhancing the accuracy of subtyping classification.

\subsubsection{VGG16 Model}
The VGG16 architecture, a 16-layer deep convolutional neural network, achieves parameter efficiency and representation capacity through its homogeneous design and progressive feature subsampling. The network employs $3 \times 3$ convolutional kernels throughout, replacing traditional large kernels to reduce computational complexity while maintaining local feature extraction capabilities through receptive field overlap. Its five-stage hierarchical structure consists of consecutive $3 \times 3$ convolutional layers followed by $2 \times 2$ max-pooling layers, progressively halving the dimensions of the feature maps. The ReLU activation function is applied subsequent to each convolutional layer to mitigate the gradient vanishing issues.

For layer $l$, the output $\mathit{y}^{(l)} \in \mathbb{R}^{H' \times W' \times C_{\text{out}}}$ is computed via:

\begin{equation}
\mathit{y}^{(l)} = \sigma\left( \sum_{p=1}^{3} \sum_{q=1}^{3} \mathit{W}^{(l)}(p,q) \ast \mathit{x}^{(l-1)} + \mathit{b}^{(l)} \right),
\end{equation}
where $\mathit{W}^{(l)} \in \mathbb{R}^{3 \times 3 \times C_{\text{in}} \times C_{\text{out}}}$ denotes the learnable convolution kernel, $\mathit{x}^{(l-1)} \in \mathbb{R}^{H \times W \times C_{\text{in}}}$ is the input feature map, and $\mathit{b}^{(l)} \in \mathbb{R}^{C_{\text{out}}}$ is initialized with Xavier initialization. This method constrains weights to have zero mean and variance $\frac{2}{3 \times 3 \times C_{\text{in}}}$, effectively addressing gradient disappearance in deep networks.

\subsubsection{BERT Model}

BERT is a transformer-based language model that captures deep contextual representations through bidirectional self-attention. It consists of multiple transformer layers, each with multi-head self-attention, feed-forward networks, and layer normalization. Given an input sequence $\mathit{x} = [x_1, x_2, ..., x_n]$, BERT represents each token as $h_i^{(0)} = x_i + p_i + s_i$, where $x_i$, $p_i$, and $s_i$ are token, positional, and segment embeddings, respectively.

Self-attention in BERT is defined as:

\begin{equation}
\text{Attention}(Q, K, V) = \text{softmax} \left(\frac{QK^T}{\sqrt{d_k}} \right) V,
\end{equation}
where $Q, K, V$ are the query, key, and value matrices, and $d_k$ is the dimension of each attention head. The final [CLS] token output, $h_{\text{CLS}}^{(L)} = f_{\text{BERT}}(x_1, x_2, ..., x_n)$, serves as the global textual representation.

BERT-base, a standard version of BERT, is pretrained on BooksCorpus~\cite{zhu2015aligning} and English Wikipedia~\cite{devlin2019bert} to learn general language representations. It adopts self-supervised learning, where the Masked Language Model (MLM) masks 15\% of input tokens and predicts them based on context, enabling BERT to learn bidirectional dependencies and improve text comprehension.

\subsection{PRMF Block}
\label{subsec:PRMFBlock}
\begin{figure*}[ht]
\centering
\includegraphics[width=1.00\textwidth]{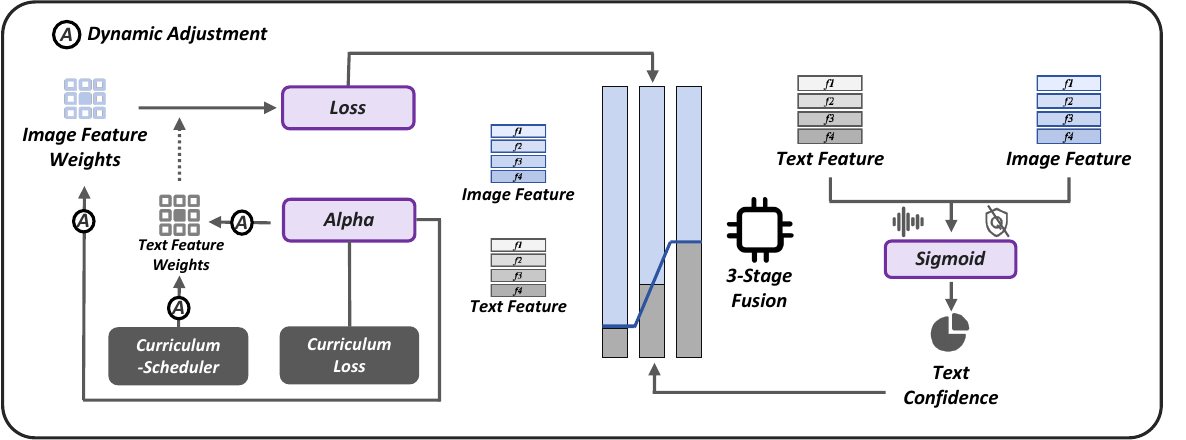}
\caption{The illustration of PRMF Block. The PRMF Block balances image and text features with a confidence-weighted fusion mechanism, reducing noisy textual impact. A curriculum learning strategy ensures a stable trainng.} 
\label{fig:prmf}
\end{figure*}

To minimize the impact of random noise in VLM-generated text descriptions, we introduce the PRMF Block as Fig.~\ref{fig:prmf}. This block employs a text confidence network that improves noise-robust through dynamic weighting. Instead of relying on simple concatenation or fixed weighting, this method adopts an adaptive text confidence mechanism and incorporates curriculum learning~\cite{hacohen2019power} to dynamically assign weights. In the context of the classification of pathological subtypes in NB, this design effectively balances early reliance on image-dominant guidance with a more effective use of text later on, thereby increasing both accuracy and reliability.

Specifically, let \(\mathit{T} \in \mathbb{R}^{d_t}\) denote the text feature vector obtained from the BERT model, and let \(\mathit{I} \in \mathbb{R}^{d_i}\) represent the global image feature vector extracted by the VGG16 model. Since the dimensions of these feature vectors typically differ (\(d_t = 768\) for BERT and \(d_i = 512\) for VGG16), a learnable linear projection is introduced to map the image features in the same dimensional space as the text features, which is defined by:

\[
\mathit{W}_{\mathit{proj}} \in \mathbb{R}^{d_t \times d_i}, \quad \mathit{b}_{\mathit{proj}} \in \mathbb{R}^{d_t},
\]
where \(\mathit{W}_{\mathit{proj}}\) is a weight matrix, and \(\mathit{b}_{\mathit{proj}}\) is a bias term. The image feature vector is then projected as:

\[
\mathit{I}' = \mathit{W}_{\mathit{proj}} \mathit{I} + \mathit{b}_{\mathit{proj}},
\]
where \(\mathit{I}' \in \mathbb{R}^{d_t}\) represents the transformed image feature vector, now aligned in the same space as the text feature vector. 

Subsequently, a small confidence network is created to dynamically assess the reliability of the text features, allowing greater reliance on the image features when the quality of the text is lacking. It takes the concatenation $[\mathit{I}; \mathit{T}]$ as input and outputs a scalar $\alpha_{\mathit{text}} \in [0,1]$:
\begin{equation}
\alpha_{\mathit{text}} = 
\sigma\Bigl(\mathit{W}_{\mathit{conf}} [\mathit{I}; \mathit{T}] + \mathit{b}_{\mathit{conf}}\Bigr),
\end{equation}
where $\mathit{W}_{\mathit{conf}} \in \mathbb{R}^{1 \times (d_i + d_t)}$, $\mathit{b}_{\mathit{conf}} \in \mathbb{R}$ are learnable parameters, and $\sigma(\cdot)$ is the sigmoid function. A larger $\alpha_{\mathit{text}}$ indicates greater textual reliability, whereas a smaller value suggests that the text may be noisy or unreliable.

Using this confidence coefficient, we adaptively balance information from both text and images. The fused feature $\mathit{F}$ is computed as follows:
\begin{equation}
\mathit{F} = 
\alpha_{\mathit{text}} \cdot \mathit{T} + 
\bigl(1 - \alpha_{\mathit{text}}\bigr) \cdot \mathit{I}',
\end{equation}
where $\mathit{F}$ is subsequently fed into the final classifier for subtype prediction. By introducing this adaptive fusion mechanism at the feature level, our MMLNB model can explicitly reduce the adverse impact of uncertain textual descriptions. This is a significant advantage over conventional multimodal methods that employ direct concatenation or fixed-weight mixing.

To maximize the effectiveness of this fusion method's noise-robust feature during training, we implement a curriculum learning strategy designed to facilitate a gradual transition from image dominance to text enhancement. In the cross-entropy loss, we introduce a weighting factor $\lambda(e) \in [0,1]$ that varies with training epoch $e$:
\begin{equation}\label{eqltotal}
\mathcal{L}_{\mathit{total}} = 
\lambda(e)\,\mathrm{CE}\bigl(\mathit{F}, y\bigr) \;+\; 
\bigl[1 - \lambda(e)\bigr] \,\mathrm{CE}\bigl(\mathit{I}, y\bigr),
\end{equation}
where $\mathrm{CE}(\cdot)$ is the cross-entropy between predicted outputs and the ground-truth label $y$, $\mathit{F}$ represents the fused feature’s prediction, and $\mathit{I}$ or its projection $\mathit{I}'$ refers to predictions based solely on the image feature. By adjusting $\lambda(e)$ to transition from 0.3 to 1.0 over time, the model initially emphasizes visual modality guidance and gradually incorporates the textual branch. Correspondingly, we freeze and unfreeze certain network components in a staged manner: early on, the textual branch is frozen to solidify the visual representations; in the middle phase, the textual branch is jointly trained; and in the final phase, the large pre-trained models are refrozen while only the fusion and classification layers are fine-tuned.

During inference, VGG16 processes the pathological image to yield $\mathit{I}$ (projected to $\mathit{I}'$), while the fine-tuned Qwen2.5-VL generates textual descriptions that are then encoded by BERT into $\mathit{T}$. These features are passed through the confidence network to determine $\alpha_{\mathit{text}}$, and the final fused feature $\mathit{F}$ (Eq.~\ref{eqltotal}) is fed into the classifier for subtype prediction.

Compared to traditional multi-modal methods, the primary advantages of our approach can be summarized as follows: (1) an explicit mechanism for dynamically adjusting text reliability at the feature level, mitigating the risk of being misled by uncertain text; (2) a curriculum-based training strategy that steadily incorporates text information, thus harnessing complementary signals without destabilizing early-stage learning; and (3) a staged freeze-unfreeze schedule tailored for large-scale pre-trained models, reducing unnecessary computations and overfitting risks.

%% file: sections/4_experiment.tex
\section{Experiment}
\subsection{Data Acquisition and Processing}
Accurate classification of NB subtypes is crucial for pathological diagnosis. However, due to the complexity of histopathological image analysis and the challenges in large-scale annotation, publicly available NB pathological image datasets remain limited. To address this, this study utilizes a private dataset from Children's Hospital, Zhejiang University School of Medicine, containing 9,000 image patches extracted from hematoxylin and eosin (H\&E)-stained whole-slide images (WSIs).

From this dataset, expert pathologists carefully selected and manually annotated 1,500 patches with textual descriptions, forming the NBITP-1.5K dataset, which is specifically designed for fine-tuning VLMs. The remaining 7,500 image patches were used to construct the NBPath-7.5K dataset, which serves as the dataset for evaluating pathological image classification models.

\begin{figure*}[ht]
    \centering \includegraphics[width=1.00\textwidth]{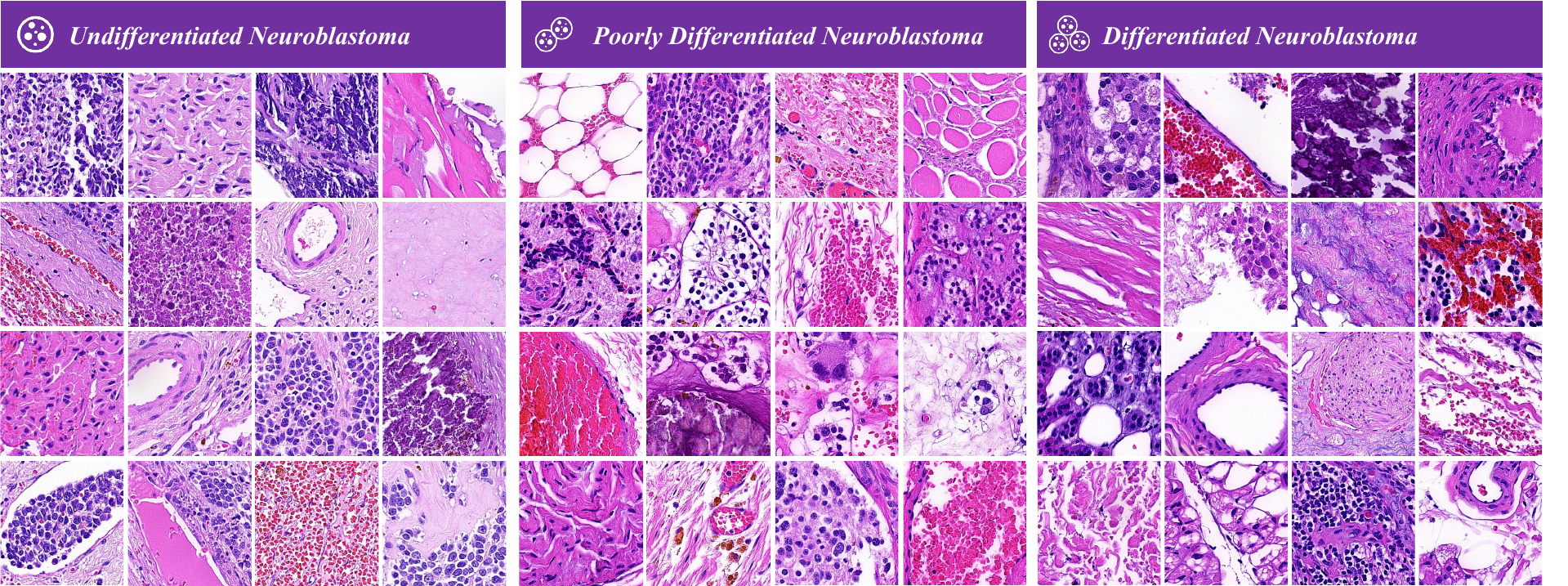} 
    \caption{Visual representation of NBPath-7.5K and NBITP-1.5K datasets. The dataset contains three subtypes of NB, namely UD, PD, and D.} 
    \label{fig4} 
\end{figure*}

The NBPath-7.5K dataset includes three NB subtypes: UD (2,200 patches), PD (2,250 patches), and D (3,050 patches). Fig.~\ref{fig4} illustrates an example of these subtypes visually. Each image patch was derived from WSIs previously diagnosed and categorized by expert pathologists. Given the extensive tissue heterogeneity within WSIs, patch-level labels were inherited from the diagnostic classification of the original WSI rather than the independent manual annotation of each patch. This labeling strategy ensures that the extracted image patches maintain the overall characteristics of the subtype while acknowledging that local histological variations may exist within each WSI.

\subsubsection{Data Processing Workflow}
\begin{figure*}[ht]
    \centering \includegraphics[width=1.00\textwidth]{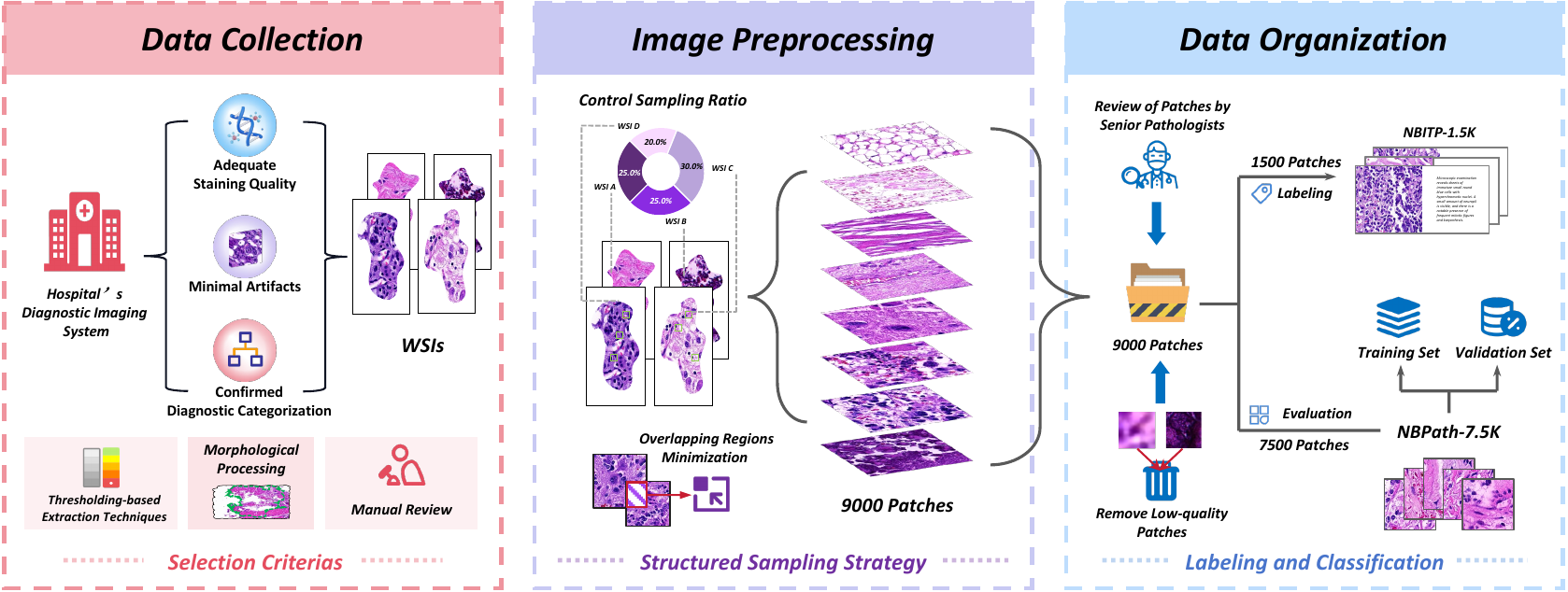} 
    \caption{The data processing workflow of the private NBPath-7.5K dataset. The process consists of three key stages: (1) Data Collection, involving the acquisition of WSIs and selection based on specific criteria; (2) Image Preprocessing, including structured sampling and patch extraction; (3) Data Organization, where images are labeled, categorized, and stored for further model training and evaluation.} 
    \label{fig:dataprocess} 
\end{figure*}

Fig.~\ref{fig:dataprocess} presents the data processing workflow, which consists of three main stages: data collection, image preprocessing, and data organization. In the data collection stage, the WSIs are obtained from clinical sources, ensuring appropriate selection criteria such as staining quality and diagnostic reliability. The Image Preprocessing stage involves structured sampling strategies that extract high-quality patches from tumor-dominant regions while preserving histopathological diversity. Finally, in the Data Organization stage, extracted patches undergo quality control, are assigned appropriate subtype labels, and are systematically stored in a structured database for subsequent experiments. This workflow ensures that the dataset is well structured, representative, and suitable for NB classification research.

\subsubsection{WSI Selection and Tumor Region Extraction}
WSIs were obtained from digitized pathology slides archived in the hospital’s diagnostic imaging system. Selection criteria included adequate staining quality, minimal artifacts, and confirmed diagnostic categorization. Slides that failed to meet these criteria were excluded to ensure consistency in image quality and diagnostic reliability.

To isolate tumor-rich regions within WSIs, an automated tissue extraction pipeline was employed. Non-tumor regions, such as stromal components and necrotic areas, were excluded using threshold-based extraction techniques. Morphological processing, including contour detection and filtering, was applied to refine the boundaries of the extracted tumor regions. The outcomes of the extraction process were subsequently examined by pathologists to ensure precision and eliminate any inaccurately identified regions.

\subsubsection{Patch Extraction and Sampling Strategy}
After extraction, patches were extracted using a structured sampling, selecting random samples from tumor-dominant areas with even spatial distribution across each WSI. Overlap was minimized to enhance morphological diversity and reduce redundancy. A controlled sampling ratio ensured a balanced distribution of patches across various cases, resulting in a dataset that accurately captures the complexity and variability of the tumor samples.

\subsubsection{Quality Control and Dataset Partitioning}
To enhance dataset reliability, extracted patches underwent a multistage quality control process. Low-quality patches, like those with blurriness or poor contrast, were removed with variance-based filtering. Contrast normalization addressed staining inconsistencies and improved feature consistency across samples. A random subset of patches was reviewed by senior pathologists to ensure the extracted regions represented the key histopathological features of each NB subtype.

After quality control, the NBPath-7.5K dataset was partitioned into training (80\%) and validation (20\%) subsets using a random split method. This ensured that the proportional distribution of NB subtypes remained consistent across both splits. Meanwhile, the NBITP-1.5K dataset was curated separately for VLM fine-tuning, leveraging detailed textual descriptions to facilitate multimodal learning. Together, these datasets provide a well-structured foundation for subsequent experimental analysis.

\subsection{Experimental Settings}

The method was implemented in Python using the PyTorch deep learning model. Experiments were conducted on a single GPU (NVIDIA RTX 4090) running Ubuntu 22.04. Each model was trained to 150 epochs, with a batch size of 32, using the Adam optimizer, and a learning rate of 0.0001.

For image pre-processing, the input images were resized to $224 \times 224$ pixels and converted to tensor format. All input images were normalized using mean and standard deviation values derived from the ImageNet dataset.

\subsection{Evaluation Indicators}
To comprehensively assess the performance of the proposed neuroblastoma subtype classification model, we employ a set of widely recognized evaluation metrics, including accuracy (Acc), balanced accuracy (BAcc), Cohen’s kappa coefficient (Kappa), weighted F1-score (F1), precision (Prec), recall (Rec), and the area under the receiver operating characteristic curve (AUROC). These indicators provide a multifaceted evaluation of the effectiveness of the model in distinguishing between histopathological subtypes. These indicators are calculated as follows:

\begin{equation}
Acc = \frac{TP + TN}{TP + TN + FP + FN},
\end{equation}
\begin{equation}
BAcc = \frac{1}{N} \sum_{i=1}^{N} \frac{TP_i}{TP_i + FN_i},
\end{equation}
\begin{equation}
Kappa = \frac{p_o - p_e}{1 - p_e},
\end{equation}
\begin{equation}
Prec = \frac{TP}{TP + FP},
\end{equation}
\begin{equation}
Rec = \frac{TP}{TP + FN},
\end{equation}
\begin{equation}
F1 = 2 \times \frac{Prec \times Rec}{Prec + Rec},
\end{equation}
\begin{equation}
AUROC = \int_{0}^{1} TPR(FPR) \, d(FPR).
\end{equation}

\subsection{Comparison Experiment}
\begin{table*}[ht]
\setlength{\abovecaptionskip}{3pt} 
\setlength{\belowcaptionskip}{6pt} 
\centering
\caption{Comparison with other advanced models on the private NBPath-7.5K dataset.}
\label{tab:comparison}
\resizebox{\textwidth}{!}{%
\begin{tabular}{@{} l *{7}{c} @{}}
\toprule
\multicolumn{1}{@{}c}{\textbf{Model}} 
& \textbf{Acc$\uparrow$} & \textbf{BAcc$\uparrow$} & \textbf{Kappa$\uparrow$} 
& \textbf{F1$\uparrow$} & \textbf{Prec$\uparrow$} & \textbf{Rec$\uparrow$} & \textbf{AUROC$\uparrow$} \\
\midrule
AlexNet~\cite{krizhevsky2012imagenet} & 74.00 & 73.98 & 60.73 & 74.07 & 74.66 & 74.00 & 88.84 \\
EfficientNet-B0~\cite{tan2019efficientnet} & 68.87 & 68.16 & 51.76 & 68.78 & 68.83 & 68.87 & 84.25 \\
ResNet-50~\cite{c24he2016deep} & 72.13 & 71.46 & 57.65 & 72.05 & 72.15 & 72.13 & 87.95 \\
ResNet-101~\cite{c24he2016deep} & 73.93 & 74.00 & 59.85 & 73.98 & 74.17 & 73.93 & 89.18 \\
VGG-16~\cite{simonyan2014very} & 71.73 & 70.50 & 57.11 & 71.50 & 72.27 & 71.73 & 87.80 \\
VGG-19~\cite{simonyan2014very} & 72.47 & 71.41 & 56.48 & 72.25 & 72.57 & 72.47 & 87.38 \\
ViT-B/16~\cite{c25dosovitskiy2020image} & 72.33 & 71.42 & 58.86 & 72.18 & 72.53 & 72.33 & 86.85 \\
DeiT-B/16~\cite{touvron2021training} & 72.60 & 72.10 & 59.13 & 72.54 & 72.56 & 72.60 & 86.90 \\
Swin-Tiny~\cite{liu2021swin} & 74.93 & 74.21 & 58.11 & 74.79 & 75.08 & 74.93 & 89.48 \\
Swin-Base~\cite{liu2021swin} & \underline{78.27} & \underline{77.67} & \underline{66.33} & \underline{78.20} & \underline{78.27} & \underline{78.27} & \underline{90.57} \\
\bottomrule
\textbf{MMLNB(Ours)} & \textbf{80.73} & \textbf{80.08} & \textbf{68.43} & \textbf{80.66} & \textbf{80.93} & \textbf{80.73} & \textbf{93.07} \\
\bottomrule
\end{tabular}
}
\end{table*}

To evaluate the effectiveness of our proposed NB subtype classification model, we compare its performance against several state-of-the-art (SOTA) deep learning models, including both CNN and transformer-based architectures. The experimental results are summarized in Table~\ref{tab:comparison}. Our model, MMLNB, achieves the highest classification Acc of 80.73\%, significantly outperforming other models. Notably, it surpasses the second best performing model, Swin-Base, by 2.46\% in Acc, demonstrating the advantage of integrating the generated textual branch. 

Additionally, the Kappa of MMLNB (68.43\%) surpasses that of other models, signifying robust agreement with expert annotations and enhanced reliability in distinguishing between subtypes of NB. The F1 (80.66\%), Prec (80.93\%), and Rec (80.73\%) further validate the capability of the model in achieving high specificity and sensitivity. Importantly, MMLNB achieves the highest AUROC score (93.07\%), confirming its strong discriminative ability to distinguish between different subtypes of NB.

Among the baseline models, Swin-Base shows competitive performance due to its hierarchical self-attention, which enhances spatial feature extraction. However, it is a unimodal model and lacks multimodal fusion capabilities, making it incapable of integrating histopathological imaging with pathology text descriptions. Transformer-based models such as ViT-B/16 and DeiT-B/16 exhibit moderate performance, but their global attention mechanisms struggle with fine-grained morphological differentiation in NB pathology. The comparatively lower performance of EfficientNet-B0 and AlexNet suggests that traditional CNN architectures, despite their efficiency, lack the representational capacity required for complex NB images.

\begin{figure*}[ht]
    \centering
    \includegraphics[width=\textwidth]{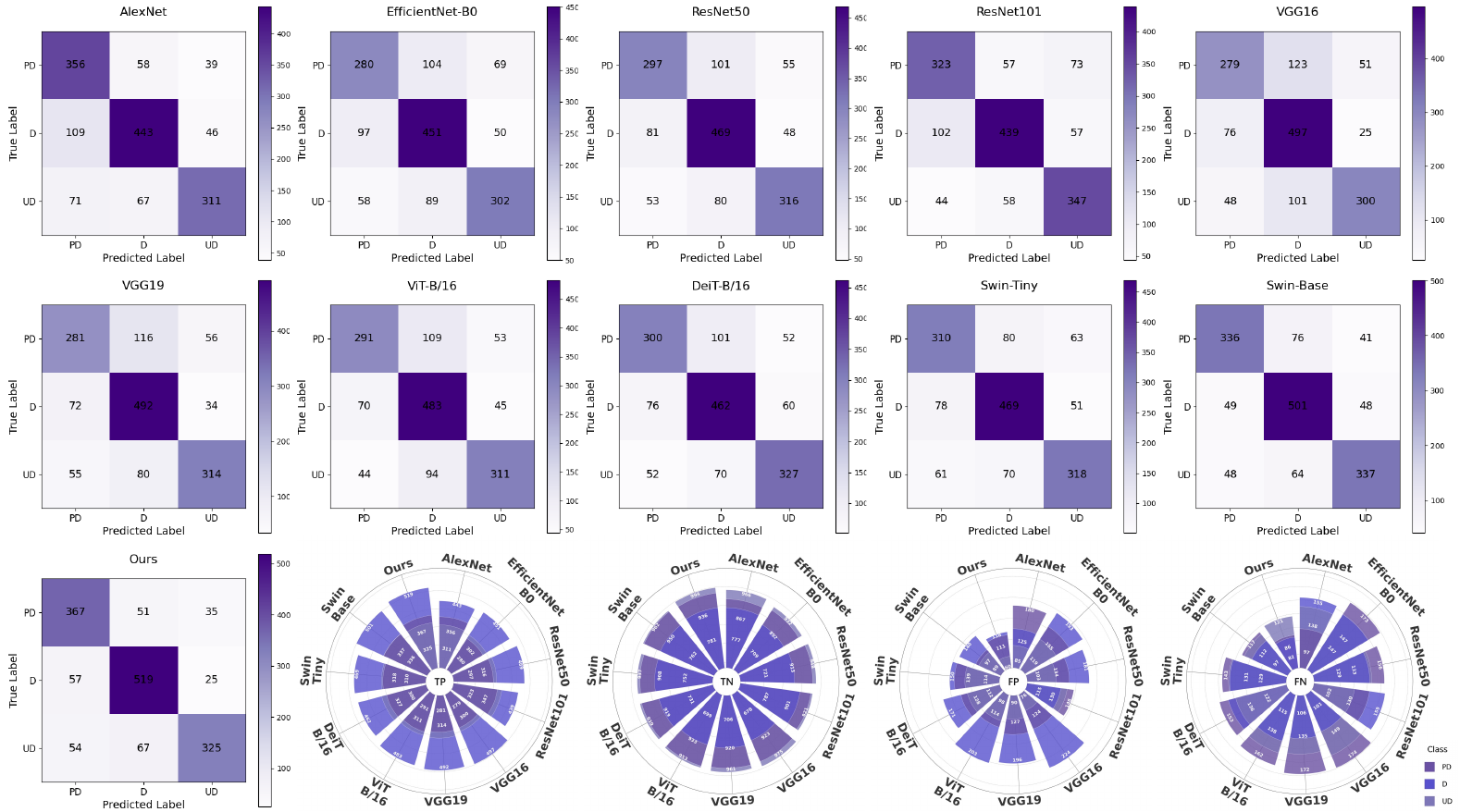}
    \caption{Visualization of experimental results from different advanced models. The subgraph contains the confusion matrix for each model and TP, TN, FP, FN, demonstrating the excellent performance of our MMLNB model.}
    \label{fig:resultvisual}
\end{figure*}

To provide a more intuitive comparison of classification performance, we visualize the experimental results of all models in Fig.~\ref{fig:resultvisual}. These visualizations illustrate the prediction performance per class, highlighting the advantages of our MMLNB model in correctly classifying different subtypes of NB. As shown, our model produces more consistent predictions with lower misclassification rates compared to other approaches. 

Overall, these results highlight the effectiveness of our proposed MML strategy, which enriches the original feature representation, leading to SOTA performance in the classification of subtypes of NB.

\subsection{Ablation Experiment}

\begin{table}[htbp]
\setlength{\abovecaptionskip}{3pt} 
\setlength{\belowcaptionskip}{6pt} 
\centering
\caption{Ablation experiments on the private NBPath-7.5K dataset.}
\label{tab:ablation}
\resizebox{\textwidth}{!}{%
\begin{tabular}{@{}l*{7}{c}@{}}
\toprule
\textbf{Model} 
& \textbf{Acc$\uparrow$} 
& \textbf{BAcc$\uparrow$} 
& \textbf{Kappa$\uparrow$} 
& \textbf{F1$\uparrow$} 
& \textbf{Prec$\uparrow$} 
& \textbf{Rec$\uparrow$} 
& \textbf{AUROC$\uparrow$} \\
\midrule
w/o Texual Branch & 71.73 & 70.50 & 57.11 & 71.50 & 72.27 & 71.73 & 87.80 \\
w/o Visual Branch & 52.37 & 68.16 & 51.76 & 68.78 & 68.83 & 68.87 & 84.25 \\
w/o Fine-Tuning VLM & 78.13 & 77.47 & 63.88 & 78.04 & 78.18 & 78.13 & 92.11 \\
w/o PRMF Block & 78.12 & 77.00 & 63.16 & 77.92 & 78.23 & 78.13 & 90.85 \\
w/o Curriculum Learning & 78.67 & 77.39 & 63.07 & 78.50 & 78.80 & 78.67 & 92.50 \\
w/o Noise Robust & 80.07 & 79.31 & 67.88 & 80.00 & 80.03 & 80.07 & 92.23 \\
\bottomrule
\textbf{MMLNB(Ours)} & \textbf{80.73} & \textbf{80.08} & \textbf{68.43} & \textbf{80.66} & \textbf{80.93} & \textbf{80.73} & \textbf{93.07} \\
\bottomrule
\end{tabular}
}
\end{table}

To assess the contributions of different components in our proposed model, we conduct an ablation study by systematically removing key elements and analyzing their impact on classification performance. The experimental results are presented in Table~\ref{tab:ablation}.

\textbf{Impact of Removing Multi-modal Branches.} Removing the textual branch reduces Acc from 80.73\% to 71.73\%, underscoring the role of pathology text in complementing histopathological imaging. Eliminating the visial branch causes a larger drop to 52.37\%, confirming the fundamental role of VGG16-extracted image features in NB subtype classification. 

\begin{figure}[ht]
    \centering  \includegraphics[width=\textwidth]{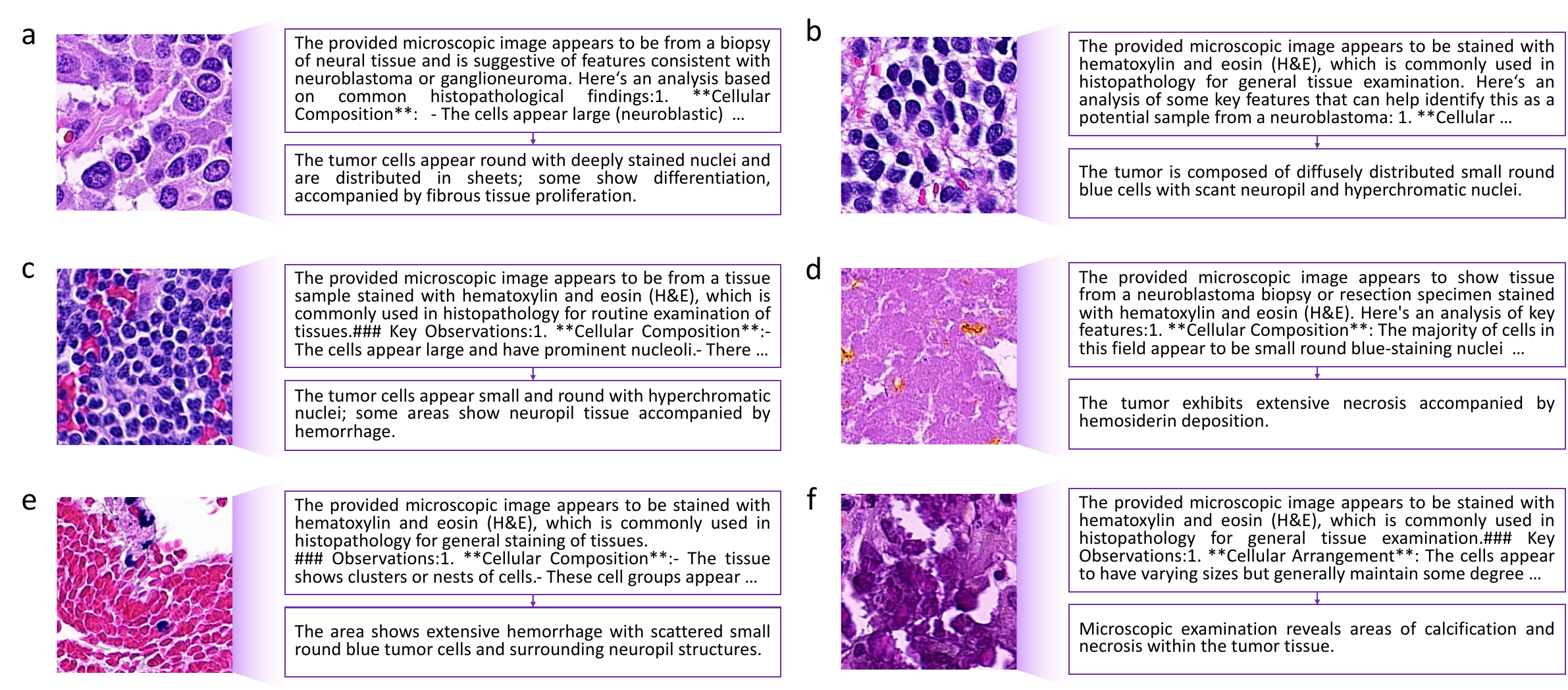}
    \caption{Comparison of generated textual descriptions before and after fine-tuning the VLM. (a) D. (b) PD. (c) UD. (d) Necrosis. (e) Hemorrhage. (f) Calcification.}
    \label{fig:textgen}
\end{figure}

\textbf{Impact of Removing Fine-tuning Stage.} Without fine-tuning Qwen2.5-VL, the model's performance decreases from 80.73\% to 78.13\%. This result indicates that LoRA-based adaptation improves the learning of domain-specific features, improving classification performance. To further illustrate the impact of fine-tuning on text generation quality, we present a visual comparison in Fig.~\ref{fig:textgen}. The figure showcases histopathological image inputs alongside their corresponding text outputs before and after fine-tuning the VLM. As observed, the fine-tuned model produces more precise textual descriptions, capturing nuanced histopathological details with greater Acc. This improvement underscores the effectiveness of LoRA-based adaptation in improving domain-specific text generation.

\textbf{Impact of Removing PRMF Block.} Removing PRMF leads to a noticeable decrease in Acc from 80.73\% to 78.12\%, and a decline in the Kappa coefficient from 68.43\% to 63.16\%. The AUROC score also drops from 93.07\% to 90.85\%. These results confirm that PRMF enables adaptive weighting between modalities and mitigates the adverse effects of noisy or uncertain textual descriptions.

\textbf{Impact of Removing the Component of PRMF Block.} When Curriculum Learning is removed, the model Acc drops to 78.67\%. This finding suggests that gradually incorporating textual features throughout training improves the model's robustness and allows for more effective utilization of multimodal information. The removal of the noise robust mechanism reduces the Acc slightly from 80.73\% to 80.07\%. Though less pronounced than other ablations, this confirms the benefit of dynamic confidence weighting in enhancing resilience to unreliable text inputs.

These ablation studies confirm the necessity of multimodal fusion, LoRA-based fine-tuning, PRMF integration, and Curriculum Learning for robust classification. Each component enhances feature extraction, interpretability, and resilience to noise, underscoring the effectiveness of our approach in advancing NB pathology classification through multimodal learning.

%% file: sections/5_conclusion.tex
\section{Conclusion}
This study presented a novel multimodal model for neuroblastoma subtype classification that significantly improved diagnostic accuracy. By integrating pathological images with text generated by VLM, we effectively tackled challenges such as domain adaptation and feature reliability. Experimental results illustrated superior classification accuracy compared to existing models. Ablation studies underscored the contributions of various components, particularly the PRMF mechanism, which increased resilience against noisy descriptions. This work highlighted the potential of multimodal AI to enhance interpretability in pathological diagnoses, providing valuable support to pathologists. Future research focused on expanding datasets to incorporate diverse variations, refining learning techniques, and exploring reinforcement learning with human feedback to improve adaptability.